# An Improved Chicken Swarm Optimization Algorithm for Handwritten Document Image Enhancement

Mugisha Stanley  
Department Of Studies  
Computer Science  
Mysore University  
India  
Email: mugishastanleys@gmail.com

Janet Lynn Tar Gutu  
Department Of Studies  
Computer Science  
Mysore University  
India  
Email: lynntar2@gmail.com

P. Nagabhushan  
Department Of Studies  
Computer Science  
Mysore University  
India  
Email:pnagabhushan@hotmail.com

*Abstract*—Chicken swarm optimization is a new meta-heuristic algorithm which mimics the foraging hierarchical behavior of chicken. In this paper, we describe the preprocessing of handwritten document by contrast enhancement while preserving detail with an improved chicken swarm optimization algorithm. The results of the algorithm are compared with other existing meta-heuristic algorithms like Cuckoo Search(CS), Firefly Algorithm(FA) and Artificial Bee Colony(ABC). The proposed algorithm considerably outperforms all the above by giving good results

*Key Words*-Chicken swarm Optimization, Hand written text, Optical Character recognition, meta-heuristic, Image enhancement

## I. Introduction

Hand written text preprocessing is a major step towards recognition of hand written text for any Optical Character Recognition (OCR)System. The quality of hand written document is degraded with time as the document ages and is affected by noise from the device such as a camera or scanner used to capture the document. Therefore recognition of the text may not be accurate by any good recognition system. The aim of preproccessing is to reduce the noise and improve accuracy of the recognition System and contrast enhancement is one of the steps of preprocessing.

A metaheuristic is a high level procedure designed to find an optimally good solution to an optimization problem with insufficient information or limited computation resources [1]. Many meta-heuristic algortihms which include Cuckoo Search(CS) [2],Ant Colony Optimization (ACO) [3],Firefly Algorithm (FF) [4],Genetic Algorithms (GA) [5] and many others have been proposed for Image enhancement . In this paper, we propose an algorithm for image enhancement of hand written text document images using an improved chicken swarm optimization algorithm.

Chicken Swarm Optimization (CSO) is a stochastic optimization algorithm proposed by mimicking the hierarchical foraging behavior of chicken in a swarm[6].

Our contributions are as follows:

1) We propose the application of the Improved Chicken Swarm Optimization in enhancement of text document Images using bi-criteria contrast enhancement which gives the user choice to adjust the contrast level while preserving detail.

The rest of this paper is organised as follows:
section 2 describes image contrast enhancement, section 3 discusses the details of the chicken swarm optimization algorithm, the proposed algorithm is presented in section 4,experimentation and discussion of results are presented in section 5, finally the conclusion in section 6.

## II. Image Enhancement

Image enhancement is described as the process of improving the visual appearance of images and make it suitable for a specific task through reduction of noise. In the spatial domain,the pixel values are controlled to attain preferred enhancement [7]. This is defined as a mapping function

$$g(m,n) = T[f(m,n)] \quad (1)$$

where $g$ is the enhanced Image, $f$ is the input image and $T$ is a function applied on $f$.

In the frequency domain, Enhancement is achieved through the following procedures:[7]
1) Compute $F(u,v)$,the Discrete Fourier Transform of the image to be enhanced,
2) multiply it by a filter function $H(u,v)$ and
3) Compute the inverse Discrete Fourier Transform to produce the enhanced image

In this paper, image enhancement in the Spatial domain was considered.

### A. Contrast enhancement using Histogram Based techniques

Given a gray scale digital Image of size $Z = M*N$ pixels, the number of possible gray levels is $L = 256$. The histogram of the image $h[n]$ can be expressed as given in [14]

$$h[n] = \{h_i \in [0, Z] | i = 0, ..., L-1\} \quad (2)$$

where $h_i$ is the frequency of the $i^{th}$ gray level in the image.

Histogram Equalization (HE) Technique is described by [13] as below
Consider an image of $M*N$ pixels. If $z_i, i = 0, 1, 2, ..., L-1$, denote the gray level range in the image. The probability density function, $p(z_k)$, of intensity level $z_k$ is given by

$$p(z_k) = \frac{n_k}{MN} \quad (3)$$

where $n_k$ is the number of times that intensity $z_k$ occurs in the image and $MN$ is the image size

A normalised histogram $p[n]$ of an image gives the approximate probability density function (PDF) of its pixel intensities. The Cumulative Distribution Function (CDF) $c[n]$, is obtained from $p[n]$. HE maps an image into the entire dynamic range $i = 0, 1, 2, ..., L-1$ using the cumulative density function mapping $T[n]$ which is given by eq(4) for an image where each pixel value is represented by 8 bits

$$T[n] = [(L-1) * \sum_{(j=0)}^{(n)} (p[j] + 0.5)] \quad (4)$$

where $n \in [0, (L-1)]$ and $p[j]$ is the normalised histogram. HE gives a flat histogram of an image which may not be exactly uniform because of the discrete nature of the pixel intensities. This results into significant change in the brightness.

*1) HISTOGRAM MODIFICATION:* However, HE suffers from large backward differences values of T[n]. This can be overcome by modifying the input histogram without compromising its contrast enhancement potential. The modified histogram can then be accumulated to map input pixel intensity values to output pixels intensity values [9]. The modified histogram can be seen as a solution of a bi-criteria optimization problem where the goal is to find a modified histogram $\widehat{h}$ which is closer to a uniformly distributed histogram $u$ but also minimize the residual difference $\widehat{h} - h_i$ where $h_i$ is the input histogram. The modified histogram would then be used to obtain the mapping function from eq.(4). This optimization problem can be modelled a weigthed sum of 2 objectives as shown in equation (5) below

$$min\|h - h_i\| + \lambda\|h - u\| \quad (5)$$

where $u \in R^{256*1}$ and $\lambda \in [0, \infty)$ is a problem parameter. if $\lambda = 0$ the Histogram obtained corresponds to the traditional HE and as $\lambda$ tends to $\infty$, it converges to the original image. Thus through varying $\lambda$ different levels of contrast can be achieved.

*2) Adjustable Histogram Equalization:* Through use of the sum of the Euclidean norm, we can obtain the analytical solution to equation(5) as follows:

$$\widehat{h} = argmin_h\|h - h_i\|_2^2 + \lambda\|h - u\|_2^2 \quad (6)$$

This gives a quadratic optimization problem

$$\widehat{h} = argmin_h[(h - h_i)^T(h - h_i) + (h - u)^T(h - u)] \quad (7)$$

The solution of above equation(7) is

$$\widehat{h} = \frac{h_i + \lambda u}{1 + \lambda} = (\frac{1}{1+\lambda})h_i + (\frac{\lambda}{1+\lambda})u \quad (8)$$

It can be observed that that the $\widehat{h}$ is a weighted average of $u$ and $h_i$. So by adjusting $\lambda$, we can easily adjust the level of enhancement.

*3) Histogram Smoothing:* The Adjustable histogram equalization procedure, as illustrated in the last section, attempts to match the original histogram to a uniform distribution and spikes occur mostly in homogeneous regions in the image. The spikes can be removed by adding a smoothness constraint to the objective and the backward difference can be used to measure the smoothness. The resultant smoothed histogram removes the gaps between intensities by matching the original histogram to a smooth distribution instead of a uniform distribution.[10] The difference matrix $D \in R^{256*256}$ is bi-diagonal with the additional term for smoothness, the optimal trade-off is obtained by

$$min\|\widehat{h} - h_i\|_2^2 + \lambda\|h - u\|_2^2 + \gamma\|Dh\|_2^2 \quad (9)$$

whose solution is a three-criterion problem

$$\widehat{h} = (1 + \lambda)I + \gamma D^T D)^{-1}(h_i + \lambda u) \quad (10)$$

where $\lambda$, the quantity for positioning the amount of contrast on a scale and $\gamma$ the amount of detail in the image to be retained.

III. THE CHICKEN SWARM OPTIMIZATION ALGORITHM

The the chicken swarm optimization algorithm(CSO) is a metaheuristic algorithm which mimicks the hierarchical foraging behavior of chicken [6].

*A. Background*

Chickens are domestic birds which live together in flocks. Their flocking habit is hierarchical in order, with each flock consisting of a dominant rooster, a number of hens and a number of chicks at the lowest level. Their exists more dominant hens which are close to the rooster and some submissive ones that stand near the chicks. A group is temporarily is disturbed when a new member enters and may disorganise the order. Stronger individuals have dominance over food access though a rooster may call members of a group it leads to eat first when food is found. This behaviour also exists in hens when they raise their chicks. This hierarchy plays an important role when it comes to foraging. There is competition for food between different chickens. But the chicks search for food around their mother. Each chick cooperates with each other[6].

*B. General Behaviour of Chicken*

1) A chicken swarm consists of many groups and each group is led by a rooster, and comprises of a number of hens and chicks.
2) Division of chicken swarm into groups and selection of the role played by each individual in a group depends on the fitness values. Chickens with best fitness values are chosen as roosters and they lead a group, and the chicken with worst values as chicks and the rest as hens. The hens find randomly which group to live in. Hens can lay eggs but not all hens would hatch their eggs simultaneously. The relationship between chicks and the hens in a group also known as mother-child relationship is randomly established.

3) The hierarchy order, domination relations and mother-child relationship will be unchanged once established, And only updates after every(G) generation.
4) Chicken in each group follow their rooster in search for food but also prevent others from stealing their own food.Chickens can randomly steal the good food already found by individuals. Each chick follows her mother in search for food. There exists competition for food and Dominant individuals always have an advantage.

It is assumed that RN, HN, CN and MN indicate the number of the roosters, the hens, the chicks and the mother hens, respectively

### C. Movement of the Chicken

Roosters are the dominant individuals in a group with the best fitness values, and therefore easily find food in a wider space. The position of the rooster is updated as follows:

$$x_{i,j}^{t+1} = x_{i,j}^t + Randn(0, \sigma^2) * x_{i,j}^t \quad (11)$$

where $x_{i,j}^{t+1}$, $x_{i,j}^t$ are the position of $j^{th}$ dimension of of the chicken $i$ at a time $t+1$ and $t$ respectively. $Randn(0, \sigma^2)$ is a Gaussian distribution with mean 0 and standard deviation $\sigma^2$ defined as shown in eq.(12) below

$$\sigma^2 = \begin{cases} 1 & \text{if } f_i \leq f_k \\ \exp\left(\frac{f_k - f_i}{|f_i| + \epsilon}\right), & otherwise \end{cases} \quad (12)$$

where $k$ is the index of the rooster $k \in [1, N]$ and $k \neq 1$, randomly selected from a group of roosters, $\epsilon$, is a number which is small enough to avoid division by 0, $f$ is the fitness value of the corresponding $x$.
The position of Hens is updated as follows:

$$x_{i,j}^{t+1} = x_{i,j}^t + S1*Rand(x_{r1,j}^t - x_{i,j}^t) + S2*Rand*(x_{r2,j}^t - x_{i,j}^t) \quad (13)$$

$$S1 = \frac{exp((f_i - f_{r1}))}{abs(f_i) + \epsilon} \quad (14)$$

$$S2 = exp((f_i - f_i)) \quad (15)$$

Where $Rand$ is a uniform random number over $[0, 1]$ $r1 \in [1, ...N]$ is an index of the rooster in the population of hen $x_i$, while $r2 \in [1, ...N]$ is an index of any other randomly chosen chicken (rooster or hen ),from the swarm. $r1 \neq r2$.
All remaining individuals are defined as chick swarm. The Position of each chick is updated as follows.

$$x_{i,j}^{t+1} = x_{i,j}^t + FL * (x_{m,j}^t - x_{i,j}^t), \quad FL \in [0, 2] \quad (16)$$

where $FL$ is a parameter which indicates that a chick would follow its mother in foraging for food, $x_{m,j}^t$ is the position index of the $i^{th}$ chick's mother.

### D. Improving Chicken Swarm Optimization

Results from experiments conducted reveal that the original CSO can easily fall into local optimum and converges prematurely for high dimensional problems[6]. To end this, an improved chicken swarm optimization was proposed by making the following observations and changes. According to eq (17),Chicks can only learn from their mothers but not the roosters.Thus,The chicks will easily fall into local optimum along with their mothers.To avoid this, the Chicks have to learn from both their mothers and the roosters in a group.Therefore the position information of the chicks is updated according to the positions of their own mother and rooster in the group using eq. (17) which is a modification for eq.(16) [8].

$$x_{i,j}^{t+1} = s * x_{i,j}^t + FL * (x_{m,j}^t - x_{i,j}^t) + F * (x_{r,j}^t - x_{i,j}^t) \quad (17)$$

where $s$ is the self-learning coefficient for the chicks , $m$ is the position of the $i^{th}$ chick's mother, $r$ is the position of the rooster. F is a learning factor, indicating that the chicks would also follow the rooster in search for food. [8].

### E. The Improved Chicken Swarm Optimization Algorithm

the improved algorithm is as follows
1) Initialize a population of N chickens and define the related parameters such as the RH,HN,CN, MN,update frequency of the chicken swarm G,maximum number of generations, etc.
2) initialize the lower bound $lu$ and upper bound $up$ using the gray levels of the image,
3) Generate a population $x = (x_1, x_2, ...x_N)$ of N chickens with random solutions.
4) Evaluate the fitness$(x_i)$ and find the best solution $x_{best}$ of the chicken swarm.
5) for t=1:itermax
6) if($t\%G == 0$) Sort the chickens' fitness values and categorise the chicken swarm into several subpopulations(Roosters,Hens and Chicks) and determine the relationship between the chicks and their mother hens in a group.
7) end if
8) For every chicken in the swarm ,update their positions depending on their category, for roosters use eq(11) and evaluate their fitness,for hens eq(13) and evaluate their fitness, for chicks use eq(17) , and and evaluate their fitness.
9) Update the personal best position $x_i^*$ and the global optimal position $x_{best}$.
10) end for
11) output the optimum value $x_{best}$

## IV. THE PROPOSED ALGORITHM

### A. Fitness Function

Image Contrast Enhancement is posed as an image histogram optimization problem where the goal to find a histogram that minimizes a cost function given in equation (9) whose solution is a three-criterion problem given by eq(10) as described by Arici in [9].

### B. Enhancement Algorithm

1) Input image a gray scale image and extract the image matrix,
2) Obtain the histogram, $h[n]$, for the image matrix.
3) Set the parameters, $\lambda$ amount of contrast, and the amount of detail to be retained $\gamma$,
4) Get a Difference matrix, D, with backward-difference of histogram, Required for histogram smoothing.
5) Apply the Improved Chicken Swarm Optimization algorithm to Obtain the optimised histogram, with parameters Maxmimal generations (iterations)$Itermax$, Population size $N$, Dimension $D$, $G$, The population size of roosters $RN$, The population size of hens $HN$, The population size of Chicks $CN$, The Population size of mother hens $MN$, input histogram $h[n]$, $\lambda$, $\gamma$ and the difference matrix, $D$.
6) Obtain the normalised histogram, p[n] from the optimised histogram using the expression below.
$p[n] = h[n]$/Number Of Pixels;
7) The Approximate Cumulative Distribution Function (CDF), $c[n]$, is obtained from $p[n]$.
8) Map back the image to spatial domain using a mapping function of the modified equation(4)
$T[n] = (\lambda + 1)255 * \sum_{(j=0)}^{(n)}(p[j] + 0.5)$
9) Then the required performance metrics are calculated.

### C. Parameter tuning and Analysis

During the experimentation the following were considered:

1) It is beneficial to keep more Hens roosters thus for any given population $RN < HN$, All hens lay eggs, but they do not all hatch simultaneously, so in a swarm $HN > MN$, Naturally, each mother hen can raise more than one chick, but it was assumed that the population of adult chickens would be greater than that of the chicks. Therefore $CN < MN$ [6].
2) The number of times the hierarchical statuses of the chicken updates(G) greatly affects the convergence rate of the ICSO algorithm. If its large, the ICSO algorithm would take a longer time to converge to a global best solution and if it was too small, the quality of the image was not good. For this experimentation, G=10 was considered to yield good results in a comparatively less time.
3) The self-learning coefficient for the chicks $s$ exponentially decreases from 0.9 to 0.4 with increase in the number of iterations. $s$ updated as shown in eq.(18)

$$s = s_{min} * (s_{max}/s_{min})^{(1/(1+10*t/Itermax))} \quad (18)$$

where $s_{min}$ is the final value of the iterations, $s_{max}$ is the initial value of the iterations, t is the current value of the iterations, $Itermax$ is the maximum number of iterations [8]
4) The Value of $\lambda$, the amount of contrast is usually varied on the scale 0-20 and the amount of detail in the image to be retained $\gamma$ is adjusted on a scale $1000 - 10^9$.

For Image 1 in fig. 1, $\lambda = 5$ and $\gamma = 50,000$ was used. However for image 2 in fig. 2, best results were obtained when $\lambda = 4$ and $\gamma = 10,000$,

5) A comparative analysis was done with other algorithms where all of the common parameters of these algorithms like the population size, dimensions, and maximum number of generations are set to be the same for a fair comparison. The Maximum number of iterations was set to 1000. The Other Parameters of the different algorithms were set as shown in Table I.

TABLE I
PARAMETER SETTINGS OF THE DIFFERENT ALGORITHMS

| Algorithm | Parameters | Values |
|---|---|---|
| ICSO | Population of chickens $N$ | 20 |
| | RN | $0.05 * N$ |
| | HN | $0.75 * N$ |
| | MN | $0.1 * HN$ |
| | CN | $N - RN - CN$ |
| | G | 10 |
| | F | 0.4 |
| | FL | $rand[0.4, 1]$ |
| Cuckoo Search | Population of Cuckoos | 20 |
| | Discovery Rate | 0.25 |
| | $\alpha$ | 1 |
| Firefly | Population of fire flies | 20 |
| | $\alpha$ | 0.5 |
| | $\beta_{min}$ | 0.2 |
| | $\gamma$ | 1 |
| ABC | Population of Bees | 20 |

## V. EXPERIMENTATION AND RESULTS

### A. Quality measures

The evaluation criteria considered here is the entropy (H)of the images, mean, variance and the (Peak Signal to Noise Ratio)PSNR. The expressions are given below.

1) The mean squared error (MSE) represents the average of the squares of the "errors" between the original image and our enhanced image. The MSE of 2 images $X(i,j)$ and $Y(i,j)$ is given by

$$MSE = \frac{\sum_{j=1}^{N}\left(\sum_{i=1}^{M}(X(i,j) - Y(i,j))^2\right)}{MN} \quad (19)$$

Peak signal to noise ratio is given in eq.(20) by the following expression: [11]

$$PSNR = 20log_{10}(\frac{255}{\sqrt{MSE}}) \quad (20)$$

2) According to Shannon's definition of the entropy ,[12] The Entropy of the was calculated using the formula below

$$H(X) = \sum_{i=1}^{n} P(x_i)I(x_i) = -\sum_{i=1}^{n} log_b P(x_i) \quad (21)$$

where b is the base of the logarithm. In this experimentation b=2.

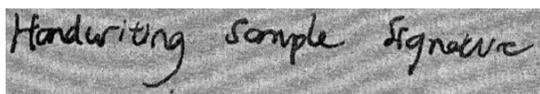

Fig. 1. Image 1, A 600x100 noisy Image of a handwritten document on poor quality paper

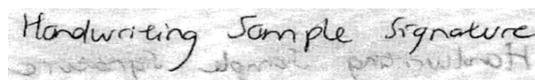

Fig. 2. Image 2, A 600x100 noisy Image of a handwritten document with writings in a background

3) Gonzalez in [13] gives the measures used to determine the mean intensity $m$ of an image according to eq(23) and Variance according to eq(24).

If $z_i, i = 0, 1, 2, ..., L-1$, denote the values of pixel intensities in an $M * N$ image. The probability, $p(z_k)$, of intensity level $z_k$ in a digital image is [13]

$$p(z_k) = \frac{n_k}{MN} \quad (22)$$

where $n_k$ is the frequency of the intensity $z_k$ in the image and $MN$ is the image size. The Mean intensity $m$ of the image is given by

$$m = \sum_{k=0}^{L-1} z_k p(z_k) = 1 \quad (23)$$

The variance of intensities, which the measure of the spread of the z values about the mean, is given by

$$\sigma^2 = \sum_{k=0}^{L-1} (z_k - m)^2 p(z_k) \quad (24)$$

*B. Results and Discussion*

The algorithm was implemented in matlab and simulations conducted on a computer with a Core i3, 2.7 GHz CPU, 2 GB of RAM installed with Octave software. The proposed Algorithm is tested on gray scale images.

The results are shown in table II for entropy, Mean, Variance and PSNR values for output images using different algorithms; Cuckoo Search(CS), FireFly(FF), Artificial Bee Colony (ABC) and Improved Chicken Swarm Optimization(ICSO). Table II shows the visual appearance for each output image. Image in fig 1 is for a hand written document text on poor quality paper, obtained using a flat bed scanner while Image 2 in fig.2 is another noisy image with some writings in the background For the results in table II, Each algorithm was executed independently 10 times respectively for each image and the average values were calculated.

It is clear from the visual results in table III that the images enhanced using the proposed ICSO show a good enhancement in terms of global quality. From the obtained numerical results from Table II, it is also obvious that the use of the ICSO algorithm out performs other techniques.

For future work, It is recommended to compare the performance of the ICSO with other new metaheuristic algorithms.

## VI. CONCLUSION

In this paper, we proposed enhancement of handwritten document images using an improved Chicken swarm optimization algorithm. There are many algorithms for image enhancement but the key challenge is maintaining uniform luminance in an image while preservation of details. The two contradict each other so they need to be balanced. hence we had to devise a strategy to increase contrast while preserving detail. We optimised the input histogram with the improved Chicken swarm algorithm while increasing the contrast and preserving details to improve recognition accuracy.

TABLE II
PSNR, VAR, MEAN AND ENTROPY MEASURES FOR EACH ALGORITHM

|  |  | Image 1 | Image 2 |
|---|---|---|---|
| Entropy | Original Image | 6.8835 | 6.3492 |
|  | FF | 2.0588 | 0.6894 |
|  | CS | 1.2734 | 0.7032 |
|  | ABC | 1.1145 | 0.4694 |
|  | **ICSO** | **0.9824** | **0.4891** |
| PSNR | FF | 24.4208 | 25.5536 |
|  | CS | 24.1181 | 25.4875 |
|  | ABC | 24.179 | 25.5545 |
|  | **ICSO** | **24.0908** | **25.4835** |
| Mean | FF | 231.4378 | 247.2733 |
|  | CS | 238.3807 | 247.6984 |
|  | ABC | 238.8735 | 247.2406 |
|  | **ICSO** | **243.3040** | **250.8344** |
| Var | FF | 63.2248 | 36.8658 |
|  | CS | 53.8975 | 34.8933 |
|  | ABC | 54.4533 | 36.8774 |
|  | **ICSO** | **44.0278** | **25.4385** |

TABLE III
VISUAL RESULTS

| | Image 1 | Image 2 |
|---|---|---|
| Original | | |
| FF | | |
| CS | | |
| ABC | | |
| ICSO | | |